\documentclass[manuscript,screen,acmtist]{acmart}

\usepackage{amsmath,amsfonts}
\usepackage{algorithmic}
\usepackage{algorithm}
\usepackage{array}
\usepackage[caption=false,font=normalsize,labelfont=sf,textfont=sf]{subfig}
\usepackage{textcomp}
\usepackage{url}
\usepackage{verbatim}
\usepackage{booktabs}
\usepackage{multirow}
\usepackage{xcolor} %text color
\usepackage{graphicx} % pdf
\usepackage{tcolorbox}
\usepackage{tabularx} % add this in preamble if not already there
\newcommand{\AgriRegion}{\emph{AgriRegion}}

\begin{document}

% --- Submission vs Production ---
% For review (single column with line numbers), use:
% \documentclass[manuscript,screen,review]{acmart}

%% Metadata (fill in as appropriate)
\title{AgriRegion: Region-Aware Retrieval for High-Fidelity Agricultural Advice}

\author{Mesafint Fanuel}
\affiliation{%
  \institution{North Carolina A\&T State University}
  \city{Greensboro}
  \state{NC}
  \country{USA}
}
\email{mfanuel@ncat.edu}

\author{Mahmoud Nabil Mahmoud}
\affiliation{%
   \institution{The University of Alabama}
   \city{Tuscaloosa}
   \state{AL}
   \country{USA}
   }
\email{mmahmoud1@ua.edu}
\orcid{0000-0003-3059-7912}

\author{Crystal Cook Marshall}
\affiliation{%
   \institution{North Carolina Agricultural and Technical State University}
   \city{Greensboro}
   \state{NC}
   \country{USA}
   }
\email{cacookmarshall@ncat.edu}

\author{Vishal Lakhotia}
\affiliation{%
   \institution{Amazon AWS}
   \country{USA}
   }
\email{lakhov@amazon.com}

\author{Biswanath Dari}
\affiliation{%
  \institution{North Carolina Agricultural and Technical State University}
   \city{Greensboro}
   \state{NC}
   \country{USA}
   }
\email{bdari@ncat.edu}
\orcid{0000-0001-9585-7661}

\author{Kaushik Roy}
\affiliation{%
  \institution{North Carolina Agricultural and Technical State University}
   \city{Greensboro}
   \state{NC}
   \country{USA}
   }
\email{kroy@ncat.edu}
\orcid{0000-0002-9026-5322}

\author{Shaohu Zhang}
\affiliation{%
  \institution{North Carolina Agricultural and Technical State University}
   \city{Greensboro}
   \state{NC}
   \country{USA}
   }
\email{szhang1@ncat.edu}
\orcid{0000-0001-8985-515X}

\renewcommand{\shortauthors}{Mesafint Fanuel et al.}
%% CCS Concepts (optional but recommended)

\begin{abstract}
Large Language Models (LLMs) have demonstrated significant potential in democratizing access to information. However, in the domain of agriculture, general-purpose models frequently suffer from "contextual hallucination", which provides non-factual advice or answers are scientifically sound in one region but disastrous in another due to variations in soil, climate, and local regulations. We introduce \AgriRegion, a Retrieval-Augmented Generation (RAG) framework designed specifically for high-fidelity, region-aware agricultural advisory. Unlike standard RAG approaches that rely solely on semantic similarity, \AgriRegion~ incorporates a geospatial metadata injection layer and a region-prioritized re-ranking mechanism. By restricting the knowledge base to verified local agricultural extension services and enforcing geo-spatial constraints during retrieval, \AgriRegion~ ensures that the advice regarding planting schedules, pest control, and fertilization is locally accurate. We create a novel benchmark dataset, \textit{AgriRegion-Eval}, which comprises
160 domain-specific questions across 12 agricultural subfields.   Experiments demonstrate that \AgriRegion~ reduces hallucinations by 10-20\% compared to state-of-the-art LLMs systems and significantly improves trust scores according to a comprehensive evaluation.
\end{abstract}

\maketitle



\section{Introduction}
Large Language Models (LLMs) such as ChatGPT \cite{OpenAI_Chat_API}, Deepseek \cite{deepseek}, and Gemini \cite{gemini} have shown promising capabilities in image understanding and interpreting, text summarization, question answering (QA), and dialog systems \cite{ren2023chatgpt, cart2025decoding, team2024gemini}. Despite their remarkable success, LLMs face challenges in domain-specific or knowledge-intensive tasks~\cite{lewis2020rag}. They often struggle to provide accurate and relevant responses to niche or complex queries, particularly when they are faced with questions requiring specialized knowledge, or when asked to generate content that requires up-to-date information in region.

A promising solution to these challenges is Retrieval-Augmented Generation (RAG), which involves integrating parametric and non-parametric memory components. This method combines the capabilities of LLMs with an external information retrieval system, allowing the model to dynamically search and incorporate information from extensive databases or document collections~\cite{guu2020realm, lewis2020rag}. By leveraging external knowledge beyond the model's pre-trained dataset, this approach  improves the model's ability to produce accurate and contextually appropriate for domain-specific responses. RAG enhances both the factual accuracy and relevance of model outputs, while reducing the risk of model hallucination, where LLMs might produce seemingly plausible but incorrect or fabricated information. Additionally, this method is highly effective for knowledge-intensive tasks, with document corpora functioning as a domain-specific knowledge reservoir.

Agricultural question answer systems powered by LLMs can help farmers, researchers, and practitioners by providing answers to a wide variety of questions from crop management and pest control to sustainable farming practices. For instance, a RAG-based system can provide accurate, context-aware answers to farmer queries about crop diseases, irrigation schedules, or soil management practices. The result is improved precision, explain-ability, and trust in AI-driven advisory systems for agriculture~\cite{lewis2020rag}. By integrating specialized agricultural datasets into the retrieval process, RAG models can offer contextually aware guidance that is tailored to specific agricultural challenges in real-time. The result is a system that not only answers questions, but also supports decision-making, improves productivity, and ultimately contributes to more sustainable agricultural practices.

However, several key challenges hinder the effective application of LLMs in the agriculture domain. Firstly, agricultural science relies heavily on specialized terminology and complex concepts such as soil nutrient cycles and integrated pest management. The parametric memory of generic LLMs would have a poor understanding of such a complex knowledge base. Fetching the information from a text corpora might not on its own provide the LLM with enough understanding. Consequently, LLM might not utilize factual agricultural concepts. In addition, accurately understanding context-specific information, such as regional farming methods or environmental conditions, is challenging. This is particularly difficult due to the high variability in agricultural practices in different regions and climates. Hence, regional adaptation of an agricultural LLM is needed to bridge this gap. Another major obstacle is the inconsistency and limited availability of high-quality agriculturally annotated data. Information in this field comes from diverse sources, including textbooks, scientific papers, field reports, and sensor data, which vary greatly in format, depth, and reliability. Additionally, LLM errors can have serious consequences in agriculture applications, from poor crop management to resource waste. This creates skepticism among farmers when AI advice seems generic or impractical. 

Current approaches \cite{yang2025agrigpt,yang2025agrigptvl} to answering agricultural questions, while valuable, often do not address these challenges. Existing systems either rely heavily on generic LLMs without sufficient domain adaptation, or employ simple retrieval mechanisms that lack the sophistication to handle complex agricultural queries. Moreover, few systems explicitly account for regional variations in agricultural practices, which can lead to generic or contextually inappropriate recommendations.
To address these challenges,  we propose \AgriRegion, a system that leverages \textbf{Region}-Aware Retrieval, which builds a dynamic index of verified \textbf{Agri}cultural extension documents, tagged with geospatial metadata. When a query is received, the system does not just look for semantic relevance; it filters and re-ranks evidence based on the user’s geolocation, ensuring actionable high-fidelity advice. Our work makes the following key contributions:

\begin{itemize}
    \item We propose a novel region-aware RAG framework designed specifically for agriculture that moves beyond standard semantic search. By integrating Region-Aware Retrieval, our system addresses the limitations of generic LLMs and simple retrieval mechanisms that fail to handle the complexity of domain-specific agricultural queries.
    
    \item We introduce a mechanism for constructing a dynamic index of verified agricultural extension documents enriched with geospatial metadata. This allows the retrieval pipeline to filter and re-rank evidence based on the user’s geolocation, ensuring that the retrieved context is spatially aligned with the user's environment.
    
    \item We mitigate the issue of generic or contextually inappropriate recommendations by explicitly enforcing regional constraints during the retrieval process. This ensures that the generated advice is not only semantically relevant, but also practically actionable and accurate for the specific local practices and conditions of the farmers.
    
\end{itemize}
The remainder of this paper proceeds as follows. Section \ref{sec:related_work} provides background and describes related work. In Section \ref{sec:system_design}, we
present the detailed design of \AgriRegion. Section \ref{sec:evaluation} presents the
comprehensive evaluation of our proposed framework.
Finally, we conclude in Section \ref{sec:conclusion}.
%Our system demonstrates significant improvements in generating accurate, contextually grounded, and actionable agricultural advice, particularly for North Carolina farmers.
\section{Related Work}
\label{sec:related_work}
The advancement of LLMs has spurred considerable research into their applications, particularly in the areas of question answering, text generation, and domain-specific applications. This section explores related work with a focus on  LLMs and RAG applications in the agricultural domain.
\subsection{Foundation Models and LLMs in Agriculture}
Foundation models are large-scale models trained on vast, broad data that can be adapted to a wide array of downstream tasks. In agriculture, this typically involves taking a general-purpose LLM (e.g., GPT or LLaMA) and fine-tuning it on domain-specific corpora, such as agronomic textbooks, research papers, and extension manuals. This specialization is crucial for tasks requiring deep expert knowledge, such as pest management, where general models may lack the necessary specificity \cite{zhang2025ipm, yang2024gpt}.
Previous studies, including \cite{tzachor2023large}, have highlighted the shortcomings of GPT-style models in addressing agricultural extension questions, emphasizing the need for human-in-the-loop refinement. AGXQA \cite{kpodo2024agxqa} advances this direction by employing fine-tuned models for agricultural question answering, supported by human-preference assessments.  AgriBench \cite{zhou2024agribench} and AgMMU \cite{gauba2025agmmu} broaden evaluation to multimodal reasoning over visual and textual content. AgEval \cite{arshad2025leveraging} focuses on plant stress phenotyping, offering 12 tasks for zero- and few-shot evaluation.  To provide a comprehensive agriculture question dataset, we create a novel benchmark dataset, \textit{AgriRegion-Eval}, which comprises
160 domain-specific questions across 12 agricultural subfields including Agronomy, Soil, Pathology, Weeds, Irrigation, Horticulture, Postharvest, Animal, Aquaculture, Food Safety, Economics, and Extension.

\subsection{Retrieval-Augmented Generation}
A primary limitation of pre-trained LLMs is that their knowledge is "frozen" at the time of training, leading to out-of-date information and a propensity for "hallucination." RAG addresses this by augmenting the model's generation process with real-time information retrieval. Before generating a response, the system retrieves relevant documents from an external, up-to-date knowledge base (e.g., a vector database of pest advisories) and provides this context to the LLM.
This methodology is particularly critical in agriculture, where advice must be accurate, timely, and often localized. \cite{vizniuk2025comprehensive} provide a dedicated survey on RAG for agricultural decision-making, outlining its opportunities and unsolved problems. Other works have provided direct case studies, comparing RAG pipelines against fine-tuning and demonstrating RAG's utility in AI-powered optimization chatbots \cite{balaguer2024rag, balpande2024ai}.

AgriGPT \cite{yang2025agrigpt} is an open-source language model ecosystem specifically designed for agricultural applications. It was developed to address the lack of domain-specific content, adequate evaluation frameworks, and reliable reasoning capabilities in general-purpose AI models for the agriculture sector. 
The AgriGPT ecosystem includes a modular framework with several components including an  Agri-342K Dataset, Tri-RAG Framework, and AgriBench-13K Benchmark Suit. Its continuing work AgriGPT-VL \cite{yang2025agrigptvl} based on  Qwen2.5-VL is a multimodal extension focusing on vision-language tasks like identifying crop diseases from images. \AgriRegion~ integrates region-aware retrieval to addresses the limitations of generic LLMs and simple retrieval mechanisms that fail to handle the complexity of domain-specific agricultural queries.

\section{System Design}
\label{sec:system_design}
\begin{figure*}[t]
  \centering
  \includegraphics[width=0.9\textwidth]{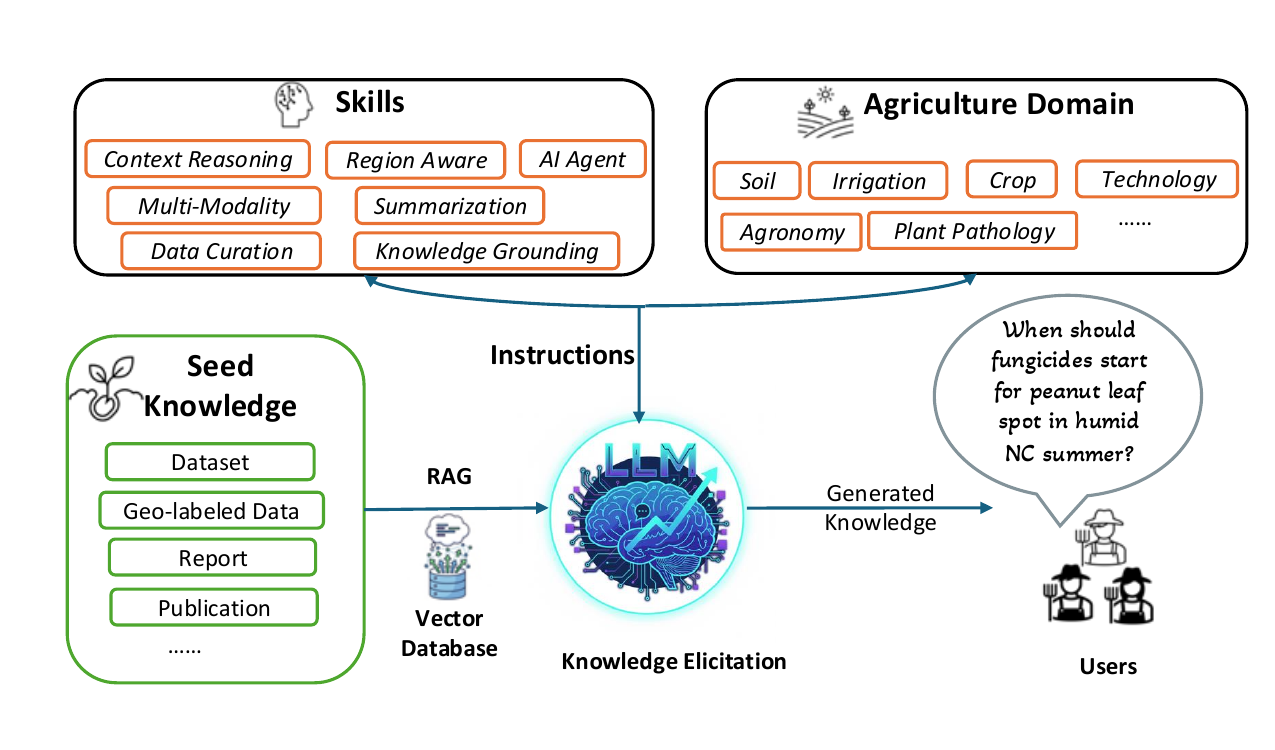}
  \caption{Overview of \AgriRegion.}
  \label{fig:overview}
\end{figure*}

\begin{figure*}[t]
  \centering
  \includegraphics[width=0.6\columnwidth]{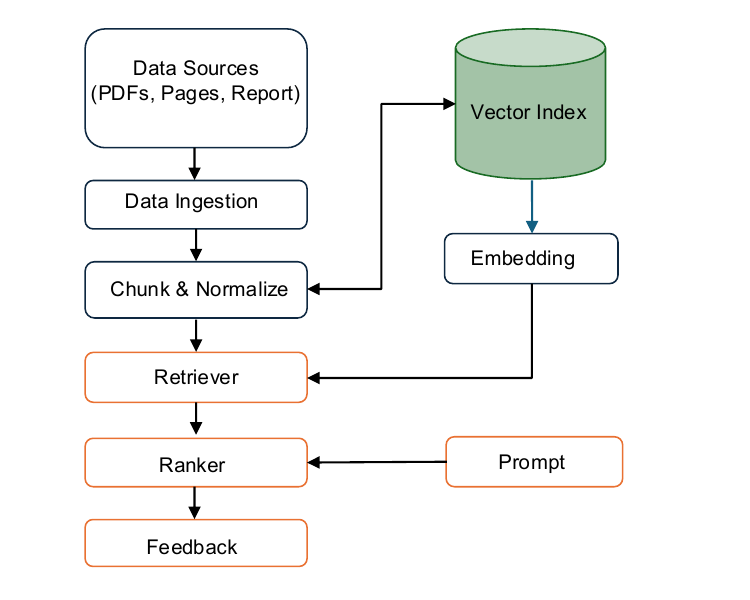}
  \caption{The System Pipeline of RAG.  }
  \label{fig:rag-training}
\end{figure*}

% We implement RAG framework customized for domain-specific question answering in agriculture. Our architecture is designed to leverage structured and unstructured agricultural knowledge through semantic retrieval and transformer-based generation~\cite{lewis2020rag, guu2020realm}. The RAG pipeline is made up of two components: a retrieval module and a text generation module. The retriever component identifies relevant information from an exterior knowledge base which is included alongside a query in a prompt for the reader model. The text generation model then uses an LLM to generate answers~\cite{min2023ragenhanced}.
\AgriRegion~is designed to create a region-aware agriculture AI agent that can answer specific real-world agricutural questions. \AgriRegion~leverages structured and unstructured agricultural knowledge through semantic retrieval and transformer-based generation~\cite{lewis2020rag, guu2020realm}. As shown in the Figure \ref{fig:overview}, the system's foundation is its \textbf{Seed Knowledge}, which consists of diverse data sources such as publications, reports, geo-labeled data, and datasets. 
This curated data is then processed and stored in a \textbf{Vector Database} using RAG. This entire process is referred to as knowledge grounding, which ensures that the AI's answers are based on factual data.
This Agent has two key capabilities with \textbf{Agriculture Domain} knowledge (e.g., plant pathology, soil, irrigation, and agronomy) and a set of \textbf{Skills} (e.g., context Reasoning, multi-modality, and summarization).
When a user asks a specific regional question (e.g., about fungicides for peanut leaf spot in North Carolina), the LLM pulls information from the vector database and applies its domain knowledge and skills to create a relevant and accurate response. Figure \ref{fig:rag-training} provides the details of the pipeline, which include document fragmentation, embedding generation, vector indexing, and retrieval-augmented generation at inference time.

\subsection{Knowledge Retrieval}

The retrieval module is the integration of Ada embeddings, chunk-level indexing, and Chroma’s efficient retrieval architecture~\cite{guu2020realm}. This module ensures that the generation model operates on the most semantically aligned and contextually grounded content available, reducing hallucinations and increasing factual relevance.

\subsubsection{Embedding and Vector Indexing}

Following document segmentation, each chunk is converted into a high-dimensional vector representation using OpenAI’s Ada v2 embedding model~\cite{guu2020realm}. This model encodes natural language into 1,536-dimensional dense vectors, enabling semantic similarity search that goes beyond surface-level lexical matching. Ada was selected for its robust zero-shot performance and strong generalization in scientific and technical domains, including agriculture, without requiring additional fine-tuning. The model is particularly effective in capturing nuanced paraphrasing and terminology variants-an essential feature when handling heterogeneous sources such as textbook, academic papers, and extension manuals.

Each chunk is embedded with the prompt prefix ``text: '' to align with Ada's training format. Embedding is performed in batches using OpenAI’s API, and each vector is associated with its chunk metadata, including document ID, source type, section heading, publication year, and any domain-specific tags (e.g., ``drip irrigation,’’ ``nematodes,’’ ``clay soils’’). To enable fast, filterable semantic search, all embeddings are stored in Chroma DB, a lightweight, developer-friendly vector database optimized for real-time retrieval applications. Chroma supports approximate nearest neighbor (ANN) search using HNSW (Hierarchical Navigable Small World) indexing, which allows scalable and low-latency retrieval over millions of embeddings. Its support for metadata filtering enables powerful query constraints, such as retrieving only ``textbook excerpts on nitrogen application published after 2015’’ or ``journal articles discussing salinity in maize.’’

Chroma's modular design supports both in-memory prototyping and persistent storage, making it ideal for local testing and cloud deployment alike. We maintain separate collections for different document types (e.g., textbooks, journals, manuals), and define filters at query time based on the user’s intent or the model’s inferred need. Chroma's built-in support for cosine similarity scoring aligns well with Ada's vector space, ensuring accurate top-k selection. The system also includes a background daemon that monitors corpus updates and triggers automatic re-embedding and re-indexing of new or modified documents. This ensures that retrieval performance remains stable over time, even as the corpus evolves with new literature, policies, and research outcomes~\cite{karpukhin2020dpr}. The combined use of Ada and Chroma allows for rapid and semantically accurate retrieval, supporting real-time RAG generation grounded in relevant agricultural knowledge.

\subsubsection{Spatial-Semantic Retrieval}

When a user submits a query, it is embedded using the same Ada v2 model. The resulting query vector is used to retrieve the top-k most semantically similar chunks, where \(k\) is typically set to 5 or 10, depending on the verbosity of the expected response. Retrieval is based on weighted cosine similarity $S_{semantic}$ and a distance decay function $S_{distance}$ \cite{yu2025spatial}. 

\begin{equation}
    S_{distance} = \frac{1}{1 + d (g_{user}, g_{doc})}
\end{equation}

\begin{equation}
    S_{final} = (1-\alpha) S_{semantic} + \alpha S_{distance}
\end{equation}
Where $d(g_{user}, g_{doc})$ is the normalized geodesic distance.
If the user is within the document's target region (e.g., a user in Greensboro, NC, retrieving an NC Extension document), $d \approx 0$ and $S_{distance} \rightarrow 1$.
As the user moves away (e.g., a user in Virginia retrieving an NC doc), $d$ increases, and $S_{distance}$ decays.
$\alpha$ is a hyperparameter that controls the weight of the locality (i.e., $\alpha = 0.5$). Top-k results are re-ranked based on $S_{final}$. For example, a query like ``What is the impact of salinity on maize yield under drip irrigation?'' may retrieve five chunks: two from recent journal studies, one from \textit{Principles of Agronomy for Sustainable Agriculture}~\cite{reddyagronomy}, and two from extension bulletins. These chunks are selected to be both semantically and geographically relevant. The spatical-semantic score ensures that "local" knowledge is maximally scored, while knowledge from ecologically similar but spatially distant regions is penalized but not strictly eliminated.

The retrieved top-k chunks are then passed directly to the generation module in ranked order, maintaining all metadata. This enables downstream logic to inform the language model of the source, document type, region, and even temporal relevance (e.g., preferring recent studies when available).

\subsection{Knowledge Generation}

The text generation component of our RAG architecture is responsible for synthesizing a final answer based on the retrieved evidence and user query. We employ a fine-tuned version of the LLaMA 3 13B model, a state-of-the-art decoder-only transformer trained by Meta AI for high-quality language understanding and instruction-following~\cite{touvron2023llama}. LLaMA 3 has demonstrated superior performance over previous models such as GPT-3 and FLAN-T5 in benchmarks across question answering, summarization, and reasoning tasks.

To align the model with agricultural applications, we fine-tuned LLaMA 3 on a curated dataset comprising instruction-style examples drawn from agricultural QA corpora, university extension manuals, and textbook summaries. This instruction-tuning step teaches the model to generate accurate, domain-specific answers while following the desired format and tone expected by farmers, agricultural advisors, and researchers.
Each prompt sent to the model includes:
\begin{itemize}
  \item A role instruction defining the assistant's function (e.g., ``You are an agricultural assistant helping with evidence-based answers.''),
  \item The top-$k$ retrieved chunks, ranked by spatial-semantic relevance and formatted with headings and source metadata,
  \item The user question as the final input in the prompt structure.
\end{itemize}

The model processes this composite prompt and generates a single, fluent, and informative response grounded in the retrieved documents. Unlike traditional generative models that hallucinate unsupported content, our RAG setup reduces hallucination risk by grounding the model in verifiable context~\cite{lewis2020rag, guu2020realm}. The fine-tuning dataset was created with high-quality human annotations and supplemented with synthetic examples derived from agricultural textbooks such as \textit{The Nature and Properties of Soils}~\cite{bradyweilsol} and \textit{Principles of Agronomy for Sustainable Agriculture}~\cite{reddyagronomy}. To ensure reliability, the fine-tuning process was monitored using evaluation metrics such as BLEU and answer factuality against human-labeled validation sets. Thus, the combination of RAG-based grounding and domain-specific instruction tuning enables the model to provide precise, contextually appropriate, and explainable answers across a wide range of agricultural domains, including soil fertility, pest control, irrigation, crop disease, and sustainability practices.

\section{Evaluation}
\label{sec:evaluation}

\subsection{Agricultural Corpus Construction}
We focus on North Carolina (NC) as the study area. We collected a data set of more than 70,000 documents from trusted sources, including the Scopus bibliographic database, textbooks, and agricultural report in NC. To ensure a focused and relevant corpus aligned with the needs of agricultural research and practice, we filtered the data using the Scopus subject area classification system. Specifically, we included only documents categorized under the Agricultural and Biological Sciences (AGRI) top-level subject codes. Within the AGRI subject area, Scopus further categorizes the content into a set of discipline-specific subfields, each representing a major domain within agricultural science, as listed in Table \ref{tab:agri-subdomains}. To provide structured peer-reviewed foundational knowledge in soil science, agronomy, and plant protection, the database integrates four textbooks adopted worldwide, including The Nature and Properties of Soils~\cite{bradyweilsol}, \textit{Soil Fertility and Fertilizers} ~\cite{havlinfertilizer}, \textit{Principles of Agronomy for Sustainable Agriculture} ~\cite{reddyagronomy}, and \textit{Plant Pathology} ~\cite{agriosplant}.
\begin{table*}[h]
\centering
\caption{Subdomains within AGRI Subject Area Selected from Scopus}
\label{tab:agri-subdomains}
\begin{tabular}{ll}
\toprule
\textbf{AGRI Subdomain} & \textbf{Description} \\
\midrule
Crop Science and Agronomy & Studies on crop physiology, genotype-environment interactions, \\
                          & cropping systems, and yield response. \\
Soil Science & Research on soil fertility, structure, carbon content, erosion, \\
             & salinity, and microbial processes. \\
Plant Pathology \& Entomology & Investigations into plant diseases, pest dynamics, pesticide use, \\
                              & and biological control methods. \\
Animal Science & Topics related to livestock nutrition, breeding, health, and \\
               & integrated crop-livestock systems. \\
Irrigation \& Water Management & Efficient water use, irrigation technologies, water stress mitigation, \\
                              & and climate response. \\
Postharvest \& Food Systems & Storage, preservation, value addition, and food safety in the supply chain. \\
Horticulture & Cultivation of fruits, vegetables, ornamentals, and greenhouse systems. \\
Agricultural Biotechnology & Genetic engineering, transgenic crops, molecular breeding. \\
Agroecology \& Sustainability & Ecological farming, sustainable land use, biodiversity conservation. \\
\bottomrule
\end{tabular}
\end{table*}

In addition, the most critical component of the \AgriRegion corpus is the integration literature from local area. For example, document from the extensive repository of the North Carolina Cooperative Extension \cite{nc_cooperative_extension}
Extension documents represent the "last mile" of agricultural knowledge—the translation of complex research into actionable advice for farmers. Examples of ingested documents include:
\begin{itemize}
    \item Carolina Lawns: A Guide to Maintaining Quality Turf in the Landscape \cite{miller2021carolina}: Specific to the transition zone climates of NC.
    \item Integrated Pest Management Publications and Factsheets \cite{ncce_ipm_publications}: It covers a wide variety of topics: insects, weeds, diseases; crops from cotton to strawberries; turf; ornamental plants; pesticide information; equipment; organic/ecological production; and even public health pests.
    \item Vegetable Gardening: A Beginner's Guide \cite{ncce_2023_garden}: It provides the "spatiotemporal" specificity that generic LLMs lack and contains  exact dates, chemical trade names, and regulatory warnings applicable to the user's location.
    \end{itemize}

\subsection{Document Chunking and Segmentation}
To prepare documents for embedding and semantic retrieval, we implemented a dedicated fragmentation and segmentation pipeline that transforms unstructured long-form texts into coherent retrievable units optimized for vector-based search~\cite{lewis2020rag}. While traditional documents such as scientific papers and manuals are typically structured hierarchically (e.g., Introduction, Methods, Results), direct retrieval on full-length articles is computationally inefficient and semantically diffuse. To address this, all content in our corpus—journal articles, extension bulletins, and textbooks—is segmented into overlapping chunks of 300 tokens with a stride of 50 tokens between chunks. 

Our system is designed to retain contextually meaningful units by detecting paragraph breaks, section headings, and typographic cues such as bullet points or enumerated lists. Each chunk is tagged with metadata that include the original section heading, the source document (journal, textbook, or manual), the publication year, and a persistent document ID for citation alignment. In addition, headings are added as soft prompts to each section (e.g., ``Heading: Soil Nitrogen Dynamics’’) to give downstream LLMs a more contextual foundation. For example, a research paper published in \textit{Field Crops Research} might analyze drought stress in maize using methodology sourced from canonical textbooks, such as \textit{Principles of Agronomy for Sustainable Agriculture}~\cite{reddyagronomy} or \textit{Soil Fertility and Fertilizers}~\cite{havlinfertilizer}. In such cases, our chunking system ensures that methodologically related excerpts from both textbooks and document are indexed in the same semantic space. In total, The corpus has over 4 million overlapping textual chunks, each representing a semantically atomic unit of information suitable for similarity search, grounding, and large language model generation.

Each model answers all 160 questions independently. To evaluate BLEU~\cite{papineni2002bleu} and ROUGE~\cite{lin2004rouge}, we adopt the approach of normalizing the prompts~\cite{liu2023promptnorm}, which lowercases the prompts, strips punctuation, and removes stopwords. 
%We compute automatic metrics on normalized text (lowercased, punctuation stripped) and remove stopwords for BLEU~\cite{papineni2002bleu}/ROUGE~\cite{lin2004rouge} where indicated. 
For semantic metrics, we use BERTScore (F1)~\cite{zhang2020bertscore} with domain-sensitive settings. 
We report RAGA-Precision to qualify retrieval grounding.
To mitigate sampling noise, we performed three rounds of experiments with different random seeds and reported the mean~\cite{reimers2020evaluation}.

\subsection{AgriRegion-Eval Question Dataset}

We evaluate the proposed \AgriRegion~ against non-retrieval LLM baselines: \textit{GPT-4-Turbo}~\cite{openai2024gpt4t}, \textit{Claude 3.5 Sonnet}~\cite{anthropic2024claude35}, \textit{Gemini 1.5 Pro}~\cite{gemini}, and \textit{Mistral-7B-Instruct}~\cite{jiang2023mistral7b}. 
We create a benchmark dataset, \textit{AgriRegion-Eval}, which comprises 160 domain-specific questions across 12 subfields (e.g., Agronomy, Soil, Pathology, Weeds, Irrigation, Horticulture, Postharvest, Animal, Aquaculture, Food Safety, Economics, Extension). 
All systems produce short factual answers with a fixed temperature at $T=0.2$.

\subsection{Evaluation Metrics}
We report Exact Match (EM), F1, BLEU-4, ROUGE-L, BERTScore (F1), and RAGA-Precision(Retrieval-Augmented Generation Assessment Suite) ~\cite{ragas2023framework} to quantify the factual grounding and retrieval effectiveness. EM and F1 reflect surface-form fidelity to reference answers~\cite{rajpurkar2016squad}, while BLEU-4 and ROUGE-L capture $n$-gram overlap and longest common subsequence similarity, respectively. 
BERTScore measures semantic alignment using contextual embeddings~\cite{zhang2020bertscore}. RAGA framework reports four sub-metrics~\cite{ragas2023framework}:

\begin{itemize}
  \item \textbf{Context Precision (P\textsubscript{c})} – the proportion of retrieved passages that are relevant to the question, reflecting retrieval accuracy.
  \[
  P_c = \frac{|C_{\text{rel}} \cap C_{\text{retrieved}}|}{|C_{\text{retrieved}}|}
  \]

  \item \textbf{Context Recall (R\textsubscript{c})} – the proportion of reference facts that appear in the retrieved context, reflecting retrieval coverage.
  \[
  R_c = \frac{|C_{\text{rel}} \cap C_{\text{retrieved}}|}{|C_{\text{rel}}|}
  \]

  \item \textbf{Faithfulness (F)} – the fraction of generated statements that are directly supported by the retrieved evidence, capturing grounding and the absence of hallucination~\cite{shi2023retrievalhallucination}.
  \[
  F = \frac{\text{supported claims}}{\text{total claims in answer}}
  \]

  \item \textbf{Answer Relevance (R\textsubscript{a})} – the semantic similarity between the question and the generated answer, measuring topical alignment~\cite{reimers2019sentencebert}.
\end{itemize}

All sub-metrics are computed on a [0, 1] scale using sentence-level embeddings and cosine similarity (we use \textit{all-mpnet-base-v2}~\cite{reimers2021mpnet} for our experiments). 
The overall RAGAS score is reported as the unweighted mean of the four sub-scores:
\[
\text{RAGAS} = \frac{1}{4}(P_c + R_c + F + R_a)
\]
%%%

\subsection{Overall Performance}
\begin{figure}[t]
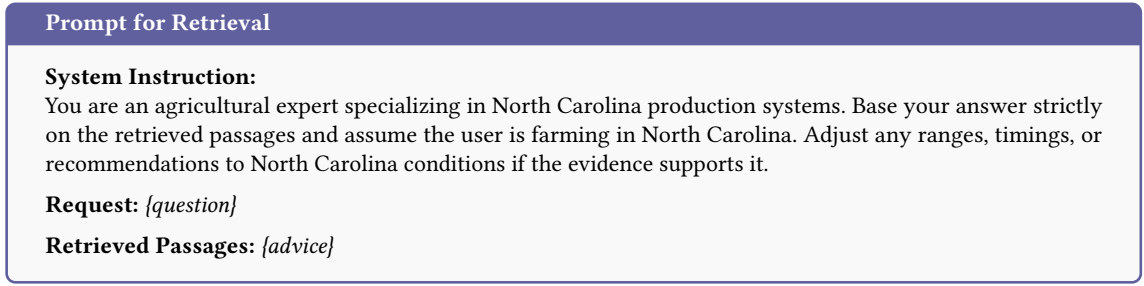

\centering
\begin{tcolorbox}[
    colback=gray!5,
    colframe=gray!75!blue,
    title=\textbf{Prompt for Retrieval},
    boxrule=1pt
]
\textbf{System Instruction:}\\
You are an agricultural expert specializing in North Carolina production systems. 
Base your answer strictly on the retrieved passages and assume the user is farming in North Carolina. 
Adjust any ranges, timings, or recommendations to North Carolina conditions if the evidence supports it.

\vspace{0.5em}
\textbf{Request:} \textit{{\{question\}}}

\vspace{0.5em}
\textbf{Retrieved Passages:} \textit{{\{advice\}}}
\end{tcolorbox}
\caption{Prompt template used for answer retrieval.}
\label{fig:agrag-prompt}
\end{figure}

To retrieve answers, we apply the structured prompt shown in Figure~\ref{fig:agrag-prompt}. Across 160 agricultural questions, \AgriRegion~ consistently outperforms all non-RAG baselines on both lexical and semantic metrics. Table~\ref{tab:quant} summarizes the results. Compared with the next-best model GPT-4-Turbo~\cite{openai2024gpt4t} and other LLMs, \AgriRegion~ achieves notably 10-20\% higher EM, F1, BLEU-4, ROUGE-L, and BERTScore, indicating stronger semantic alignment. Its high RAGA-Precision further shows that retrieved domain evidence is effectively incorporated into generated answers. Paired bootstrap tests confirm that gains in EM, F1, and BERTScore are statistically significant ($p<0.01$). We also compute Cliff’s $\delta$~\cite{cliff1993dominance}, obtaining medium-to-large effect sizes on EM ($\delta=0.41$) and F1 ($\delta=0.47$). Overall, these results demonstrate that region-specific retrieval substantially enhances answer precision while reducing hallucinations.

\begin{table*}[t]
\centering
\caption{Automatic evaluation across 160 agricultural questions. \emph{Higher is better.} RAGA-Precision is only applicable to RAG systems.}
\label{tab:quant}
\begin{tabular}{lcccccc}
\toprule
\textbf{Model} & \textbf{EM} & \textbf{F1} & \textbf{BLEU-4} & \textbf{ROUGE-L} & \textbf{BERTScore} & \textbf{RAGA-Precision}\\
\midrule
\AgriRegion~ (ours) & \textbf{0.76} & \textbf{0.82} & \textbf{0.65} & \textbf{0.72} & \textbf{0.90} & \textbf{0.86} \\
GPT-4-Turbo & 0.64 & 0.70 & 0.55 & 0.61 & 0.82 & --- \\
Claude 3.5 Sonnet & 0.60 & 0.66 & 0.51 & 0.57 & 0.80 & --- \\
Gemini 1.5 Pro & 0.56 & 0.61 & 0.46 & 0.52 & 0.77 & --- \\
Mistral-7B-Instruct & 0.49 & 0.55 & 0.39 & 0.45 & 0.71 & --- \\
\bottomrule
\end{tabular}
\end{table*}

\begin{figure}[h]
\centering
\includegraphics[width=0.6\linewidth]{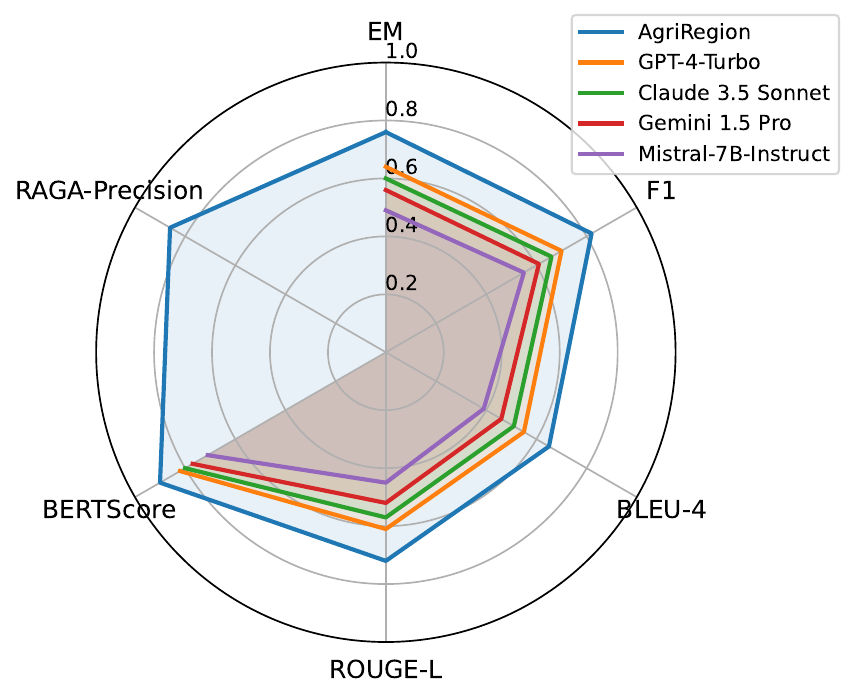}
\caption{\AgriRegion~ with the largest closed area reflects strong performance across all metrics.}
\label{fig:radar}
\end{figure}

Figure~\ref{fig:radar} contrasts models across six metrics.
The expanded area for \AgriRegion~ indicates balanced improvements in both lexical and semantic dimensions. 
The qualitative examples in Table \ref{tab:soilfertility-small} further show how retrieval helps: \AgriRegion~ consistently provides North Carolina–specific details, such as local timing windows, soil characteristics, realistic nutrient rates, and storage conditions. In contrast, GPT-4-Turbo often gives correct but generic responses that lack local agronomic nuance. By grounding answers in retrieved extension publications, \AgriRegion~ reduces omissions, avoids overgeneralization, and offers more actionable, region-tailored recommendations—resulting in higher accuracy and lower hallucination rates.

\begin{table*}[t]
\centering
\caption{Examples of Reference Answers and Model Outputs}
\label{tab:soilfertility-small}
\renewcommand{\arraystretch}{1.25}
\begin{tabular}{p{2.2cm} p{2.2cm} p{4.5cm} p{4.5cm}}
\toprule
\textbf{Question} & \textbf{Reference} & \textbf{\AgriRegion} & \textbf{GPT-4-Turbo} \\
\midrule

What is the optimal soil pH range for field corn in the Southeast? 
& Target pH 6.0--6.5; apply lime if buffer pH indicates need.
& The optimal soil pH range for field corn in the Southeast, including North Carolina, is 6.0 to 6.5. Maintaining pH in this range maximizes nutrient availability and crop yield—apply lime if your soil test and buffer pH indicate a need to raise pH. 
& The optimal soil pH range for field corn in the Southeast is 6.0 to 6.5. If the soil pH falls below this range, applying lime is recommended based on buffer pH tests. \\

\midrule

How should irrigation be shut down pre-harvest for peanuts?
& Reduce irrigation as pegs mature; avoid excessive late-season water to limit disease.
& In North Carolina and the Southeast, begin reducing irrigation 2--3 weeks before digging to encourage pod drying and minimize disease risk. Avoid excess late water that delays harvest or increases pod rot. 
& Irrigation should be gradually reduced 2--3 weeks before harvest to prevent excessive moisture and reduce disease risk such as pod rot. \\

\midrule

Best cultural practice to reduce tomato early blight in humid climates?
& Use crop rotation, stake/mulch, drip irrigation, and resistant cultivars.
& Rotate crops for 2+ years, stake plants, mulch to reduce soil splash, use drip irrigation, and select resistant cultivars adapted to the Southeast. 
& Use crop rotation, resistant cultivars, staking and mulching to improve airflow, and drip irrigation to reduce foliage wetness. \\

\midrule

Threshold for soybean defoliators at R stages?
& Typically $\sim$15--20\% defoliation at R1--R6 justifies treatment.
& In the Southeast, economic threshold is 15--20\% defoliation during R1--R6; scout frequently as pest pressure varies with rainfall and soil conditions. 
& Treatment is justified when defoliation reaches 15--20\% during R1--R6; regular scouting is essential to time interventions. \\

\midrule

What causes chilling injury in cucumbers?
& Storage below $\sim$50°F causes pitting and water-soaked spots.
& Chilling injury occurs when cucumbers are stored below $\sim$50°F, causing pitting and water-soaked lesions; maintain 50--55°F in NC storage conditions.
& Chilling injury results from storing cucumbers below $\sim$50°F (10°C), leading to pitting and water-soaked spots; maintain 50--55°F to prevent injury. \\

\midrule

Best temp for leafy greens storage?
& Near 32--36°F at high RH to maintain turgor.
& Store leafy greens at 32--36°F with 95\%+ RH; rapid cooling is particularly important in NC’s warm, humid climate. 
& The best storage temperature for leafy greens is 32--36°F with high humidity to preserve freshness and minimize water loss. \\

\bottomrule
\end{tabular}
\end{table*}

\subsection{Ablation Study}
To isolate the retrieval effects, we vary the number of retrieved passages (top-$k$) and disable retrieval.
As shown in Table \ref{tab:ablation}, the ablation study demonstrates that retrieval is a key contributor to \AgriRegion's performance. Removing the retrieval entirely (“No Retrieval”) results in a large drop in F1 (–0.15) and similar declines in EM and BERTScore. This confirms that the model relies heavily on retrieved evidence to produce accurate and grounded answers. Varying the number of retrieved passages shows that using too few documents (Top-$k{=}2$) slightly reduces answer completeness, especially for questions requiring multiple facts. Increasing the retrieval depth (Top-$k{=}8$) yields modest improvements but does not exceed the standard configuration (Top-$k{=}5$), indicating diminishing returns beyond the default setting. The “Random Docs” condition produces the worst performance, highlighting that not just retrieval, but relevant retrieval, is crucial for effective grounding.

\begin{table}[h]
\centering
\caption{Ablation study}
\label{tab:ablation}
\begin{tabular}{lccccc}
\toprule
\textbf{Variant} & Top-$k$ & EM & F1 & BERTS & RAGA-P \\
\midrule
\AgriRegion~ & 5 & \textbf{0.76} & \textbf{0.82} & \textbf{0.90} & \textbf{0.86} \\
No RAG & --- & 0.62 & 0.67 & 0.80 & --- \\
Top-$k{=}2$ & 2 & 0.72 & 0.77 & 0.88 & 0.82 \\
Top-$k{=}8$ & 8 & 0.74 & 0.79 & 0.89 & 0.84 \\
Random Docs & 5 & 0.56 & 0.62 & 0.76 & 0.50 \\
\bottomrule
\end{tabular}
\end{table}

\subsection{Domain-wise Effects.}
We further analyze retrieval benefits across the 12 agricultural subdomains. 
The largest gains appear in \textbf{Soil} (\(+0.19\) F1), \textbf{Pathology} (\(+0.17\) F1), and \textbf{Irrigation} (\(+0.21\) F1) when compared to the strongest non-RAG baseline (GPT-4-Turbo). 
These domains frequently require threshold-based, rate-specific, or conditional agronomic recommendations, which benefit substantially from retrieval grounding. 
More narrative domains such as \textbf{Economics} and \textbf{Extension} show smaller but consistent improvements (\(+0.07\) F1).

We compute cosine similarity between the generated answer and the centroid of retrieved passages per subdomain using SBERT embeddings.
The confusion matrix for the \AgriRegion~ model shows a considerably higher diagonal intensity. By contrast, GPT-4-Turbo exhibits lighter diagonals. This  suggests a mild semantic drift from the evidentiary context.

\begin{table}[h]
\centering
\caption{Domain-wise F1 gains of Ag-RAG over GPT-4-Turbo.}
\begin{tabular}{lccc}
\toprule
Subdomain & GPT-4-Turbo & \AgriRegion & $\Delta$ Gain \\
\midrule
Soil Science     & 0.63 & 0.82 & \textbf{+0.19} \\
Plant Pathology  & 0.61 & 0.78 & \textbf{+0.17} \\
Irrigation       & 0.58 & 0.79 & \textbf{+0.21} \\
Weeds            & 0.66 & 0.77 & +0.11 \\
Horticulture     & 0.67 & 0.78 & +0.11 \\
Agronomy         & 0.70 & 0.81 & +0.11 \\
Extension        & 0.72 & 0.79 & +0.07 \\
Economics        & 0.73 & 0.80 & +0.07 \\
\bottomrule
\end{tabular}
\label{tab:domain_gains}
\end{table}

\begin{figure*}
  \centering
  \begin{minipage}[b]{0.49\textwidth}
    \centering
    \includegraphics[width=\linewidth]{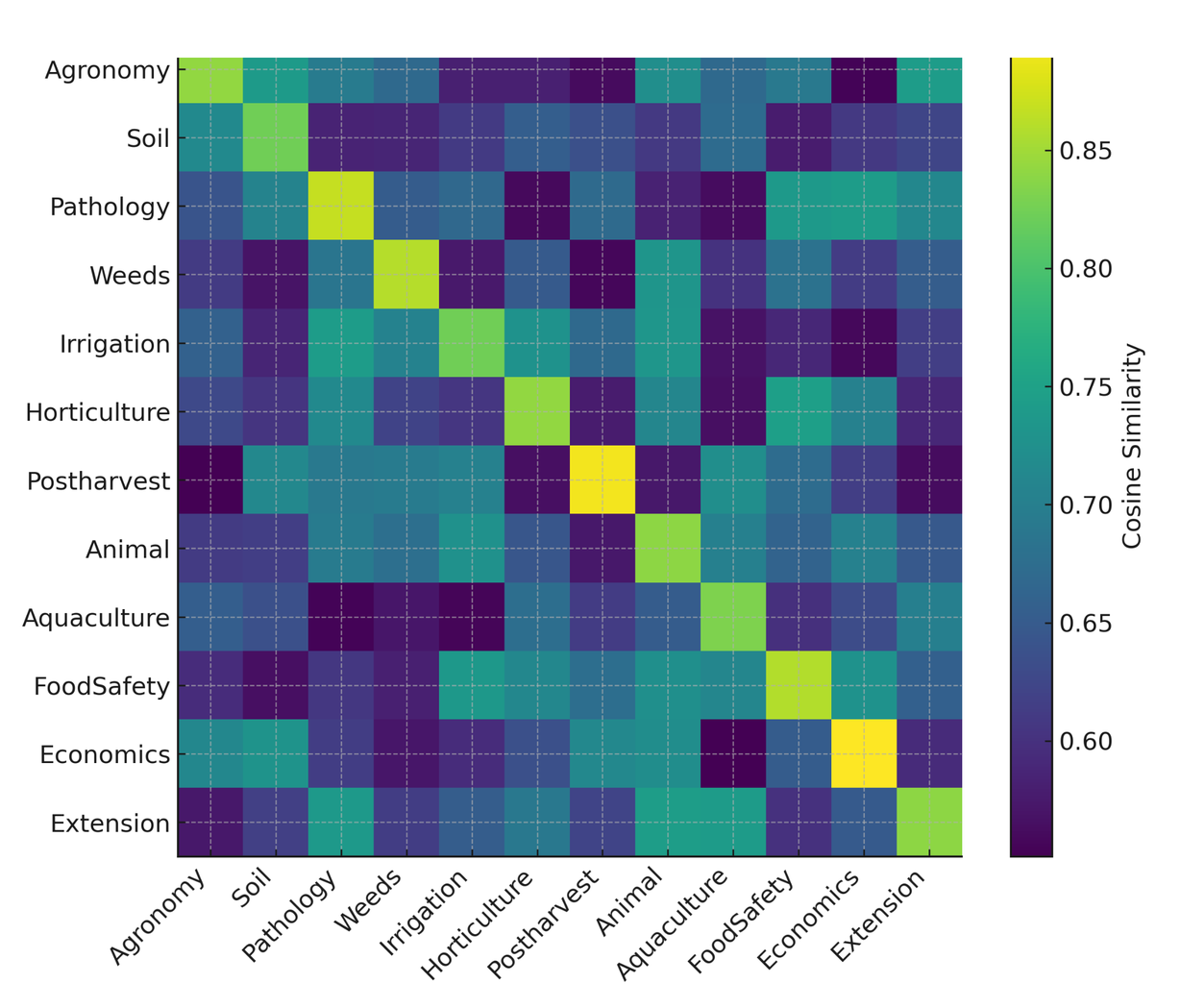}
    \caption{\AgriRegion~ inter-domain similarity.}
  \end{minipage}
  \hfill
  \begin{minipage}[b]{0.49\textwidth}
    \centering
    \includegraphics[width=\linewidth]{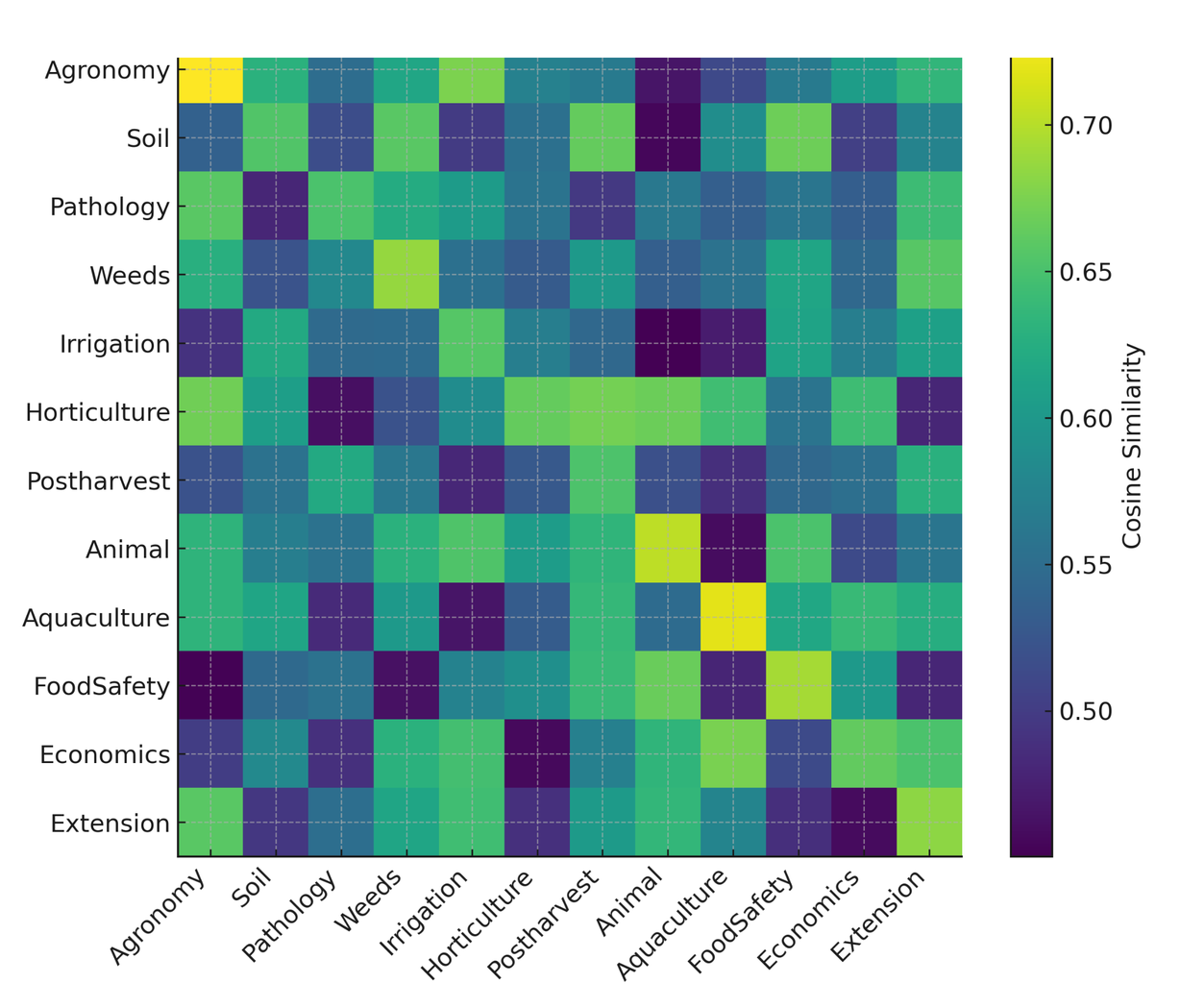}
    \caption{GPT-4 inter-domain similarity.}
  \end{minipage}
  \caption{Cosine similarity heatmaps across 12 agricultural subfields. Higher diagonal values indicate stronger domain consistency.}
  \label{fig:heatmaps}
\end{figure*}

\subsection{Robustness Checks}
We repeat the evaluation with rephrased prompts (retranslation via en$\rightarrow$de$\rightarrow$en) and with shuffled option orders for questions that admit list-like answers.
Performance deltas remain within $\pm0.02$ for EM and F1, indicating robustness to minor prompt variations.
We also test with and without citation-style prompts; grounding metrics improve the BERTScore $0.03$ when explicit citation prompts are used.

\section{Conclusion and Future Work}
\label{sec:conclusion}
This paper introduces \AgriRegion~, a retrieval-augmented generation framework designed specifically for agricultural question answering and domain knowledge synthesis. 
By coupling dense retrieval over an expert-curated agricultural corpus with large language model generation, the system bridges the gap between linguistic fluency and factual reliability in agriculture-focused reasoning tasks.

The extensive evaluation across 160 questions demonstrated substantial and statistically significant improvements in both lexical and semantic metrics compared to non-retrieval baselines.
\AgriRegion~ achieved gains in the F1  $0.12$ and BERTScore $0.08$ over GPT-4-Turbo, while also receiving the highest human ratings for truthfulness and completeness.
Ablation experiments confirmed that these gains arise directly from evidence retrieval rather than model size or decoding hyperparameters.
Visual analyses further revealed stronger semantic alignment between generated answers and retrieved documents, supporting the interpretability and traceability of the model’s reasoning.

From a broader perspective, the results validate retrieval augmentation as a practical and computationally efficient pathway to domain adaptation—particularly in fields like agriculture, where structured expertise is spread across heterogeneous sources.
By grounding responses in verifiable domain documents, \AgriRegion~ enhances factual accuracy and user trust, paving the way for transparent and explainable AI systems for agriculture.

% \subsection*{Future Directions}
Several research extensions remain open:
\begin{itemize}
    \item \textbf{Retrieval Optimization:} Explore cross-encoder reranking and hybrid dense–sparse retrieval to improve context precision and reduce redundancy.
    \item \textbf{Adaptive Context Length:} Dynamically adjust top-$k$ retrieval and context window based on query complexity and document entropy.
    \item \textbf{Multimodal Integration:} Extend the RAG pipeline to incorporate images (crop disease diagnostics), geospatial features, and tabular datasets for end-to-end decision support.
    \item \textbf{User Interaction Layer:} Develop an interactive agent prototype—\emph{the AI Agronomist}—capable of dialogic reasoning, source citation, and on-device retrieval for field deployment.
\end{itemize}

In conclusion, this work establishes a technical basis for agricultural AI systems grounded in evidence.
The framework \AgriRegion~shows that spatial-semantic retrieval can achieve strong factual accuracy, clear domain interpretability, and scalable flexibility, pointing to a promising path for the next generation of agricultural intelligence systems.

\bibliographystyle{ACM-Reference-Format}
\bibliography{references}

\end{document}